\documentclass{article}

\usepackage{microtype}
\usepackage{graphicx}
\usepackage{subfigure}
\usepackage{booktabs} 

\usepackage{tablefootnote}
\usepackage[para,online,flushleft]{threeparttable}
\usepackage{footnote}

\usepackage{hyperref}

\usepackage[accepted]{icml2020}

\icmltitlerunning{Hybrid Session-based News Recommendation using Recurrent Neural Networks}

\begin{document}

\twocolumn[
\icmltitle{Hybrid Session-based News Recommendation using \\Recurrent Neural Networks}



\icmlsetsymbol{equal}{*}

\begin{icmlauthorlist}
\icmlauthor{Gabriel de Souza P. Moreira}{nvidia,ita}
\icmlauthor{Dietmar Jannach}{uni_klag}
\icmlauthor{Adilson Marques da Cunha}{ita}
\end{icmlauthorlist}

\icmlaffiliation{nvidia}{NVIDIA - 2788 San Tomas Expressway, Santa Clara, CA, 95051.}
\icmlaffiliation{ita}{Department of Electrical Engineering and Computing, Instituto Tecnol\'ogico de Aeron\'autica (ITA) - S\~ao Jos\'e dos Campos, S\~ao Paulo, Brazil.}
\icmlaffiliation{uni_klag}{Department of Applied Informatics, University of Klagenfurt, Austria}

\icmlcorrespondingauthor{Gabriel de Souza P. Moreira}{gspmoreira@gmail.com}

\icmlkeywords{Artificial Neural Networks, Context-Aware Recommender Systems, Hybrid Recommender Systems, News Recommender Systems, Session-based Recommendation}

\vskip 0.3in
]



\printAffiliationsAndNotice{ }  

\begin{abstract}

We describe a hybrid meta-architecture -- the CHAMELEON -- for session-based news recommendation that is able to leverage a variety of information types using Recurrent Neural Networks. We evaluated our approach on two public datasets, using a temporal evaluation protocol that simulates the dynamics of a news portal in a realistic way. Our results confirm the benefits of modeling the sequence of session clicks with RNNs and leveraging side information about users and articles, resulting in significantly higher recommendation accuracy and catalog coverage than other session-based algorithms.

\end{abstract}

\section{Introduction}

Recommender systems help users to deal with information overload by providing tailored item suggestions to them. One of the earliest application domains is the recommendation of online \textit{news} \cite{karimi2018news}. News recommendation is sometimes considered as being particularly difficult, as it has a number of distinctive characteristics \cite{Zheng:2018:DDR:3178876.3185994}. Among other challenges, news recommenders have to deal with a constant stream of news articles being published, which at the same time can become outdated very quickly. Another challenge is that the system often cannot rely on long-term user preference profiles. Typically, most users are not logged in and their short-term reading interests must be estimated from only a few logged interactions, leading to a \textit{session-based recommendation problem} \cite{QuadranaetalCSUR2018}.

In recent years, we observed an increased interest in the problem of session-based recommendation, where the task is to recommend relevant items given an ongoing user session. Recurrent Neural Networks (RNN) represent a natural choice for sequence prediction tasks, as they can learn models from sequential data. \textit{GRU4Rec} \cite{hidasi2016} was one of the first neural session-based recommendation techniques, and a number of other approaches were proposed in recent years that rely on deep learning architectures, as in \cite{Liu2018stamp, Li2017narm}.

However, as shown in \cite{jannach2017recurrent,ludewig2018evaluation,LudewigMauro2019}, neural approaches that only rely on logged item interactions have certain limitations and they can, depending on the experimental setting, be outperformed by simpler approaches based, e.g., nearest-neighbor techniques.
Differently from previous works, we therefore leverage multiple types of side information with RNNs, including textual article embeddings, as well as the context of users and articles. Furthermore, we propose a meta-architecture to address the aforementioned challenges of recommending in the news domain.

\section{Technical Contribution}
Our approach is based upon CHAMELEON \cite{moreira2018chameleon, moreira2018news, moreira2019inra, moreira2019contextual}, which is a Deep Learning Meta-Architecture for News Recommendation. It supports session-based news recommendation scenarios, modeling the sequence of user clicks using Recurrent Neural Networks. The resulting system is a hybrid recommender system, which addresses the permanent user and item cold-start problem in the news domain by leveraging the textual content of news articles, the article context (e.g., recent popularity and recency) and the user context (e.g., time, location, device, previous session clicks).

\begin{figure}[h!t]
	\centering\includegraphics[width=1.0\linewidth]{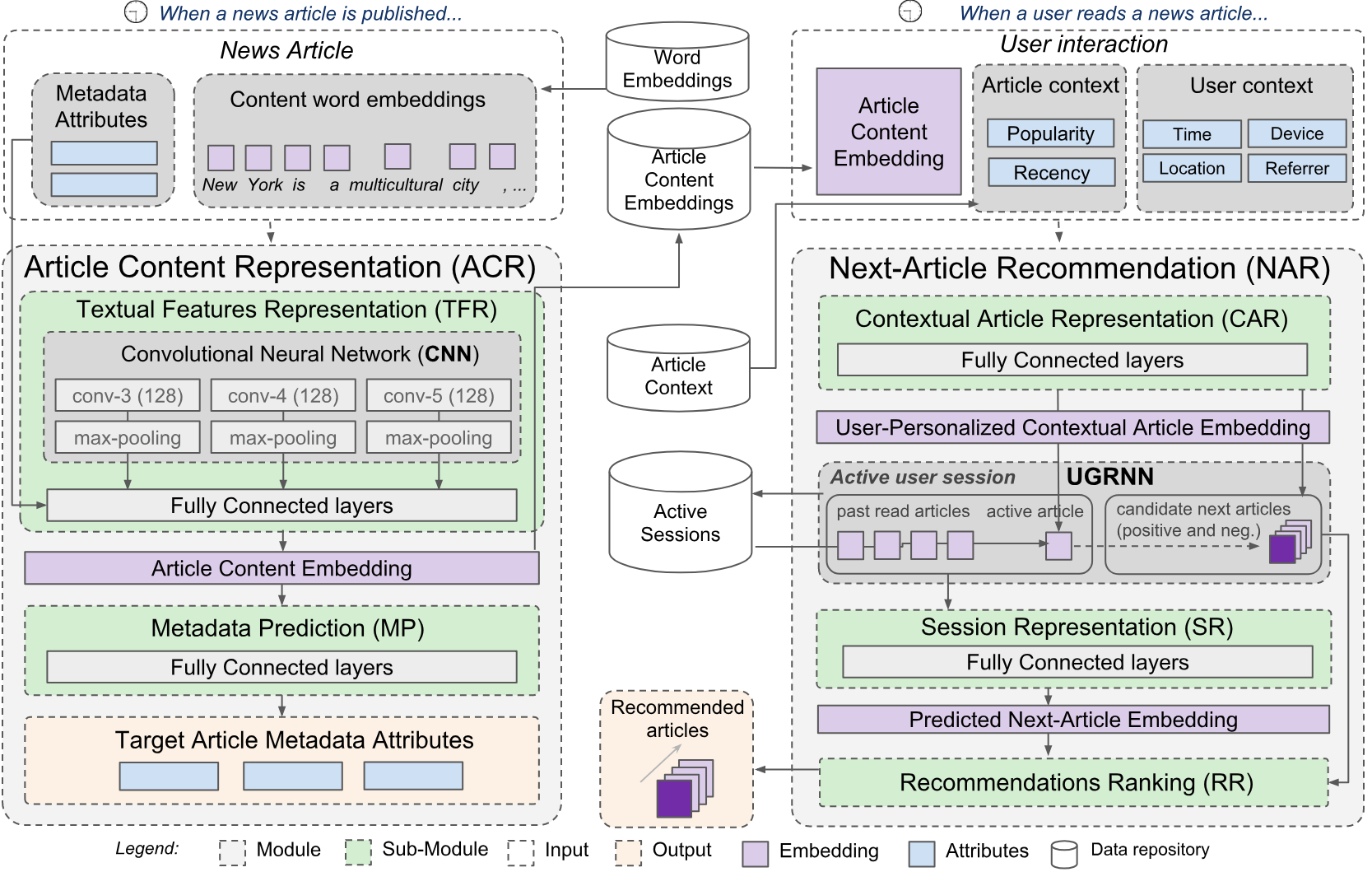}	\caption{An architecture instantiation of \textit{CHAMELEON}}
	\label{figure:chameleon_instantiation}
\end{figure}

Figure~\ref{figure:chameleon_instantiation} shows our instantiation of the \emph{CHAMELEON} framework with its two main modules:  the \emph{ACR} module on the left creates distributed representations of articles' textual content. The \emph{NAR} module on the right is responsible to generate next-click predictions. The \emph{NAR} module is trained on a ranking loss function based on similarities, which is designed to recommend fresh articles without retraining. As proposed for the \textit{DSSM} loss function \cite{huang2013learning}, it is trained to maximize the likelihood of correctly predicting the next clicked article given a user session.

\section{Evaluation Protocol}

The evaluation was performed as follows: (1) Recommenders are continuously trained on users' sessions ordered by time and grouped by hours. Each five hours, the recommenders are evaluated on sessions from the next hour; (2) For each session in the evaluation set, we incrementally revealed one click after the other to the recommender; and (3) For each click to be predicted, we created a set containing 50 negative samples articles (not clicked by the user in her session) and compute top-N metrics about accuracy, item coverage, and novelty.

\begin{figure}[h!t]
	\centering\includegraphics[width=0.99\linewidth]{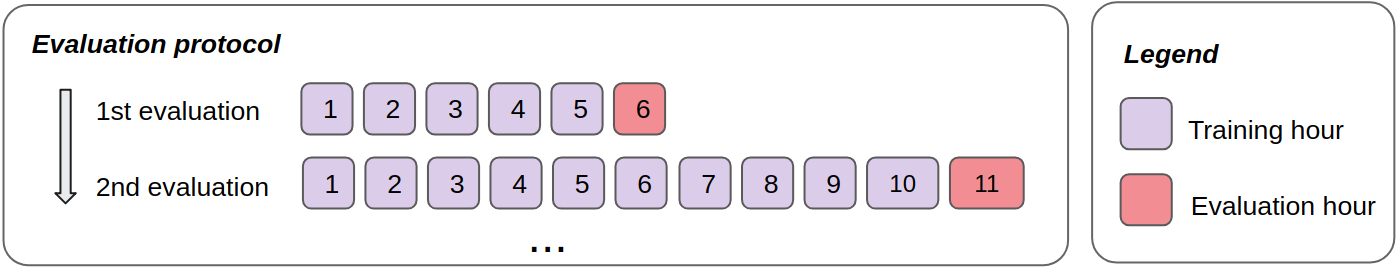}	\caption{Illustration of the evaluation protocol. After training for 5 hours, we evaluate using the sessions of the next hour.}
	\label{figure:eval_protocol}
\end{figure}

Experiments were performed with two public real-world datasets from the \emph{G1} \cite{moreira2018news} and \emph{Adressa} \cite{gulla2017adressa} news portals, described in Table~\ref{tab:datasets}.

We evaluated the following recommendation quality factors for the top-N ranked items: accuracy -- Hit Rate (\emph{HR@n}) and Mean Reciprocal Rank (MRR@n); item coverage -- (\emph{COV}) (i.e., the number of distinct articles that appeared in any top-N list divided by the number of recommendable articles); and novelty -- \emph{ESI-R}, which is based on item popularity, returning higher values when recommending long-tail items.

As baseline algorithms for session-based recommendation, we have used: two neural approaches (\emph{GRU4Rec} \cite{hidasi2016} and \emph{SR-GNN} \cite{wu2019session}); association rules-based methods (\emph{Co-Occurrence (CO)} and \emph{Sequential Rules (SR)} \cite{ludewig2018evaluation}); neighborhood-based methods (\emph{Item-kNN} \cite{hidasi2016} and \emph{Vector Multiplication Session-Based kNN (V-SkNN)} \cite{jannach2017recurrent}); and two other classical methods (\emph{Recently Popular (RP)} \cite{ludmann2017recommending} and \emph{Content-Based (CB)}.

\begin{table}[h!t]
\centering
\footnotesize
\caption{Statistics of the datasets used for the experiments.}
\label{tab:datasets}
\vspace{10pt}
\begin{tabular}{p{3.6cm}rr}
\hline
 & \textit{Globo.com (G1)}
 & \textit{Adressa} \\ \hline
Language  & Portuguese & Norwegian  \\ 
Period (days)  & 16 & 16 \\  
\# users   & 322,897 & 314,661  \\ 
\# sessions & 1,048,594 & 982,210 \\  
\# clicks  & 2,988,181 & 2,648,999 \\  
\# articles   & 46,033 & 13,820  \\ 

Avg. Sessions length (clicks)  & 2.84  & 2.70  \\
\hline
\end{tabular}

\end{table}

\section{Results}

The evaluation results are presented in Table~\ref{tab:metrics_results}, as originally reported in \cite{moreira2019contextual}. The best results for a metric are printed in bold face and marked with * if they are significantly different \footnote{As errors around the reported averages were normally distributed, we used paired Student's t-tests with Bonferroni correction at $p<0.001$ for significance tests.} from all other algorithms.

\begin{table}[h!t]
\centering
\caption{Evaluation of recommendation quality factors}
\label{tab:metrics_results}
\footnotesize
\begin{tabular}{p{1.8cm}p{1.0cm}p{1.0cm}p{1.1cm}p{1.4cm}}
\hline
 \textit{Recommender}
 & \textit{HR@10}
 & \textit{MRR@10}
 & \textit{COV@10}
 & \textit{ESI-R@10}
 \\
\hline
\multicolumn{5}{p{4cm}}{\textbf{G1 dataset}}\\ \hline
\textit{CHAMELEON} & \textbf{0.6738}* & \textbf{0.3458}* & 0.6373 & 6.4177   \\
\textit{SR} & 0.5900 & 0.2889 & 0.2763 & 5.9747 \\
\textit{Item-kNN} & 0.5707 & 0.2801 & 0.3913 & 6.5909   \\
\textit{CO} & 0.5689 & 0.2626 & 0.2499 & 5.5728 \\
\textit{V-SkNN} & 0.5467 & 0.2494 & 0.1355 & 5.1760  \\
\textit{SR-GNN} & 0.5144 & 0.2467 & 0.3196 & 5.4280  \\
\textit{GRU4Rec} & 0.4669 & 0.2092 & 0.6333 & 5.2332   \\
\textit{RP} & 0.4577 & 0.1993 & 0.0218 & 4.4904  \\
\textit{CB} & 0.3643 & 0.1676 & \textbf{0.6774} & \textbf{8.1531}*  \\
\hline
\multicolumn{5}{p{4cm}}{\textbf{Adressa dataset}}\\ \hline
\textit{CHAMELEON} & \textbf{0.7018}* & \textbf{0.3421}* & 0.7926 & 5.3410  \\
\textit{SR} & 0.6288 & 0.3022 & 0.4604 & 5.4443  \\
\textit{Item-kNN} & 0.6179 & 0.2819 & 0.5314 & 5.4675  \\
\textit{CO}  & 0.6131 & 0.2768 & 0.4220 & 5.0789  \\
\textit{V-SkNN} & 0.6140 & 0.2723 & 0.1997 & 4.6018 \\
\textit{SR-GNN} & 0.6122 & 0.2991 & 0.5197 & 5.1013  \\
\textit{GRU4Rec} & 0.4958 & 0.2200 & 0.5143 & 5.0571   \\
\textit{RP} & 0.5648 & 0.2481 & 0.0542 & 4.1465  \\
\textit{CB}  & 0.3307 & 0.1253 & \textbf{0.8875}* & \textbf{7.6715}* \\
\hline
\end{tabular}
\end{table}

\section{Conclusion}

CHAMELEON was specifically designed to address news recommendation challenges such as (a) the short lifetime of the recommendable items and (b) the lack of longer-term preference profiles of the users.
In the extensive experiments performed, CHAMELEON was able to provide recommendations with much higher accuracy than all other evaluated algorithms, and it led to the second best item coverage.


\bibliography{references}

\begin{thebibliography}{17}
\providecommand{\natexlab}[1]{#1}
\providecommand{\url}[1]{\texttt{#1}}
\expandafter\ifx\csname urlstyle\endcsname\relax
  \providecommand{\doi}[1]{doi: #1}\else
  \providecommand{\doi}{doi: \begingroup \urlstyle{rm}\Url}\fi

\bibitem[Gulla et~al.(2017)Gulla, Zhang, Liu, {\"O}zg{\"o}bek, and
  Su]{gulla2017adressa}
Gulla, J.~A., Zhang, L., Liu, P., {\"O}zg{\"o}bek, {\"O}., and Su, X.
\newblock The adressa dataset for news recommendation.
\newblock In \emph{Proceedings of the International Conference on Web
  Intelligence (WI'17)}, pp.\  1042--1048, 2017.

\bibitem[Hidasi et~al.(2016)Hidasi, Karatzoglou, Baltrunas, and
  Tikk]{hidasi2016}
Hidasi, B., Karatzoglou, A., Baltrunas, L., and Tikk, D.
\newblock Session-based recommendations with recurrent neural networks.
\newblock In \emph{Proceedings of Fourth International Conference on Learning
  Representations (ICLR'16)}, 2016.

\bibitem[Huang et~al.(2013)Huang, He, Gao, Deng, Acero, and
  Heck]{huang2013learning}
Huang, P.-S., He, X., Gao, J., Deng, L., Acero, A., and Heck, L.
\newblock Learning deep structured semantic models for web search using
  clickthrough data.
\newblock In \emph{Proceedings of the 22nd ACM International Conference on
  Conference on Information \& Knowledge Management}, pp.\  2333--2338, 2013.

\bibitem[Jannach \& Ludewig(2017)Jannach and Ludewig]{jannach2017recurrent}
Jannach, D. and Ludewig, M.
\newblock When recurrent neural networks meet the neighborhood for
  session-based recommendation.
\newblock In \emph{Proceedings of the Eleventh ACM Conference on Recommender
  Systems (RecSys'17)}, pp.\  306--310, 2017.

\bibitem[Karimi et~al.(2018)Karimi, Jannach, and Jugovac]{karimi2018news}
Karimi, M., Jannach, D., and Jugovac, M.
\newblock News recommender systems--survey and roads ahead.
\newblock \emph{Information Processing \& Management}, 54\penalty0
  (6):\penalty0 1203--1227, 2018.

\bibitem[Li et~al.(2017)Li, Ren, Chen, Ren, Lian, and Ma]{Li2017narm}
Li, J., Ren, P., Chen, Z., Ren, Z., Lian, T., and Ma, J.
\newblock Neural attentive session-based recommendation.
\newblock In \emph{Proceedings of the 2017 ACM on Conference on Information and
  Knowledge Management}, CIKM '17, pp.\  1419--1428, 2017.

\bibitem[Liu et~al.(2018)Liu, Zeng, Mokhosi, and Zhang]{Liu2018stamp}
Liu, Q., Zeng, Y., Mokhosi, R., and Zhang, H.
\newblock {STAMP:} short-term attention/memory priority model for session-based
  recommendation.
\newblock In \emph{Proceedings of the 24th {ACM} {SIGKDD} International
  Conference on Knowledge Discovery {\&} Data Mining, {KDD'18}}, pp.\
  1831--1839, 2018.

\bibitem[Ludewig \& Jannach(2018)Ludewig and Jannach]{ludewig2018evaluation}
Ludewig, M. and Jannach, D.
\newblock Evaluation of session-based recommendation algorithms.
\newblock \emph{User-Modeling and User-Adapted Interaction}, 28\penalty0
  (4--5):\penalty0 331--390, 2018.

\bibitem[Ludewig et~al.(2019)Ludewig, Mauro, Latifi, and
  Jannach]{LudewigMauro2019}
Ludewig, M., Mauro, N., Latifi, S., and Jannach, D.
\newblock Performance comparison of neural and non-neural approaches to
  session-based recommendation.
\newblock In \emph{Proceedings of the 2019 ACM Conference on Recommender
  Systems (RecSys 2019)}, 2019.

\bibitem[Ludmann(2017)]{ludmann2017recommending}
Ludmann, C.~A.
\newblock Recommending news articles in the clef news recommendation evaluation
  lab with the data stream management system odysseus.
\newblock In \emph{Working notes of the Conference and Labs of the Evaluation
  Forum (CLEF'17)}, 2017.

\bibitem[Moreira(2018)]{moreira2018chameleon}
Moreira, G. d. S.~P.
\newblock Chameleon: A deep learning meta-architecture for news recommender
  systems.
\newblock In \emph{Proceedings of the Doctoral Symposium at 12th ACM Conference
  on Recommender Systems (RecSys'18)}, pp.\  578--583, 2018.

\bibitem[Moreira et~al.(2018)Moreira, Ferreira, and Cunha]{moreira2018news}
Moreira, G. d. S.~P., Ferreira, F., and Cunha, A. M.~d.
\newblock News session-based recommendations using deep neural networks.
\newblock In \emph{Proceedings of the 3rd Workshop on Deep Learning for
  Recommender Systems (DLRS) at ACM RecSys'18}, pp.\  15--23, 2018.

\bibitem[Moreira et~al.(2019{\natexlab{a}})Moreira, Jannach, and
  da~Cunha]{moreira2019contextual}
Moreira, G. D. S.~P., Jannach, D., and da~Cunha, A.~M.
\newblock Contextual hybrid session-based news recommendation with recurrent
  neural networks.
\newblock \emph{IEEE Access}, 7:\penalty0 169185--169203, 12
  2019{\natexlab{a}}.
\newblock ISSN 2169-3536.
\newblock \doi{10.1109/ACCESS.2019.2954957}.

\bibitem[Moreira et~al.(2019{\natexlab{b}})Moreira, Jannach, and
  da~Cunha]{moreira2019inra}
Moreira, G. D. S.~P., Jannach, D., and da~Cunha, A.~M.
\newblock On the importance of news content representation in hybrid neural
  session-based recommender systems.
\newblock In \emph{Proceedings of the 7th International Workshop on News
  Recommendation and Analytics (INRA 2019) @ RecSys 2019}, volume Vol-2554.
  CEUR-WS, 09 2019{\natexlab{b}}.

\bibitem[Quadrana et~al.(2018)Quadrana, Cremonesi, and
  Jannach]{QuadranaetalCSUR2018}
Quadrana, M., Cremonesi, P., and Jannach, D.
\newblock Sequence-aware recommender systems.
\newblock \emph{ACM Computing Surveys (CSUR)}, 51\penalty0 (4):\penalty0 66,
  2018.

\bibitem[Wu et~al.(2019)Wu, Tang, Zhu, Wang, Xie, and Tan]{wu2019session}
Wu, S., Tang, Y., Zhu, Y., Wang, L., Xie, X., and Tan, T.
\newblock Session-based recommendation with graph neural networks.
\newblock In \emph{Proceedings of the AAAI Conference on Artificial
  Intelligence}, volume~33, pp.\  346--353, 2019.

\bibitem[Zheng et~al.(2018)Zheng, Zhang, Zheng, Xiang, Yuan, Xie, and
  Li]{Zheng:2018:DDR:3178876.3185994}
Zheng, G., Zhang, F., Zheng, Z., Xiang, Y., Yuan, N.~J., Xie, X., and Li, Z.
\newblock Drn: A deep reinforcement learning framework for news recommendation.
\newblock In \emph{Proceedings of the 2018 World Wide Web Conference}, WWW '18,
  pp.\  167--176, 2018.

\end{thebibliography}
\bibliographystyle{icml2020}

\end{document}